\pdfoutput=1

\documentclass[11pt]{article}

\usepackage[final]{acl}

\usepackage{times}
\usepackage{latexsym}

\usepackage[T1]{fontenc}

\usepackage[utf8]{inputenc}

\usepackage{microtype}

\usepackage{inconsolata}

\usepackage{graphicx}
\usepackage{booktabs}
\usepackage{multirow}
\usepackage{makecell}
\usepackage{amsmath}
\usepackage{amssymb}
\usepackage[most]{tcolorbox}
\usepackage{arydshln}

\NewDocumentCommand{\hongru}
{ mO{} }{\textcolor{blue}{\textsuperscript{\textit{Hongru}}\textsf{\textbf{\small[#1]}}}}

\NewDocumentCommand{\xqc}
{ mO{} }{{#1}}

\usepackage{xspace}
\newcommand{\framework}{TInR-U\xspace}

\title{TInR: Exploring Tool-Internalized Reasoning in Large Language Models}

\author{Qiancheng Xu$^{1}$, Yongqi Li$^{1\dagger}$, Fan Liu$^{2}$, Hongru Wang$^{3}$, Min Yang$^{4}$, Wenjie Li$^{1}$ \\
$^{1}$ The Hong Kong Polytechnic University \quad
$^{2}$ Southeast University \quad
$^{3}$ University of Edinburgh \\
$^{4}$ Shenzhen Institutes of Advanced Technology, Chinese Academy of Sciences \\
\texttt{qiancheng.xu@connect.polyu.hk} \quad
\texttt{liyongqi0@gmail.com} \quad
\texttt{cswjli@comp.polyu.edu.hk}
}

\begin{document}

\maketitle
\begingroup\def\thefootnote{$\dagger$}\footnotetext{Corresponding author.}\endgroup

\begin{abstract}
Tool-Integrated Reasoning (TIR) has emerged as a promising direction by extending Large Language Models' (LLMs) capabilities with external tools during reasoning.
Existing TIR methods typically rely on external tool documentation 
during reasoning. However, this leads to tool mastery difficulty, tool size constraints, and inference inefficiency.
To mitigate these issues, we explore Tool-Internalized Reasoning (TInR), aiming at facilitating reasoning with tool knowledge internalized into LLMs. 
Achieving this goal presents notable requirements, including tool internalization and tool-reasoning coordination.
To address them, we propose \framework, a tool-internalized reasoning framework for unified reasoning and tool usage. \framework is trained through a three-phase pipeline: 1) tool internalization with a bidirectional knowledge alignment strategy; 2) supervised fine-tuning warm-up using high-quality reasoning annotations, and 3) reinforcement learning with TInR-specific rewards. 
We comprehensively evaluate our method across in-domain and out-of-domain settings. Experiment results show that \framework achieves superior performance in both settings, highlighting its effectiveness and efficiency.
Codes are available at \href{https://github.com/travis-xu/TInR}{https://github.com/travis-xu/TInR}.

\end{abstract}

\section{Introduction}
Large language models (LLMs), such as Deepseek-R1~\citep{guo2025deepseek}, have demonstrated remarkable reasoning and problem-solving capabilities in complex tasks such as code generation, logical deduction, and workflow planning.
However, they often struggle in scenarios beyond their capabilities such as knowledge updates, weather inquiries, or restaurant reservations.
To address this issue, Tool-Integrated Reasoning (TIR) has been proposed, enabling LLMs to leverage external tools during the reasoning process and thus extend their capabilities beyond purely language-based reasoning to a broader range of practical applications.

\begin{figure}[!t]
    \centering
    \centerline{\includegraphics[width=1.1\columnwidth]{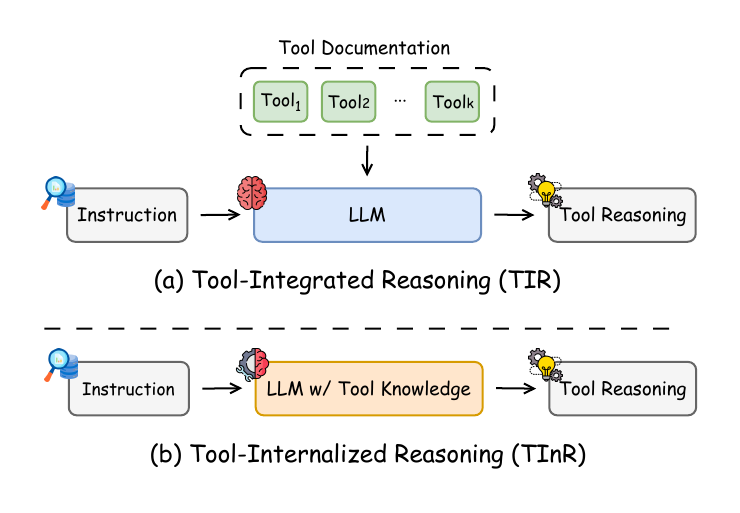}}
    \caption{Comparison between (a) Tool-Integrated Reasoning (TIR) and (b) Tool-Internalized Reasoning (TInR). TInR internalizes tool knowledge into LLMs to facilitate reasoning.}
    \label{fig:intro}
\vspace{-1em}
\end{figure}

A typical TIR process, as illustrated in Figure~\ref{fig:intro}(a), begins with a user instruction and a set of available tools accompanied by their documentation; the LLM is then expected to reason among the candidate tools to address the given instruction.
Under this setting, early TIR methods often rely on supervised fine-tuning (SFT) to teach LLMs with annotated reasoning paths, but suffer from limited generalization and adaptability~\cite{chu2025sft}. 
Thereafter, reinforcement learning (RL) 
approaches have been proposed to allow LLMs to integrate reasoning with tool learning through outcome-based feedback~\citep{qian2025toolrl,wang2025actingreasoningmoreteaching}, thus fostering more strategic and exploratory reasoning abilities. 


Despite recent progress, existing TIR methods still rely on prompt-based tool documentation to inform LLMs about available tools. In other words, tool knowledge remains external to LLM and must be explicitly provided for reasoning.
This brings several limitations:
1) \textbf{Tool mastery difficulty}. 
Tool documentation is often heterogeneous and inconsistent, making it difficult for LLMs to quickly grasp tool knowledge on the fly~\citep{yuan-etal-2025-easytool,qu2025from}.
This brings a gap between external tool knowledge and LLM’s internal understanding, which hinders effective tool mastery during reasoning.
2) \textbf{Tool size constraints}. 
As the number of tools increases, it becomes infeasible to include all tool documentation within the context window. While retrieval strategies can partially alleviate this, they introduce additional pipeline complexity and cause a potential misalignment between retrieval and tool usage~\citep{xu-etal-2024-enhancing-tool}. 
3) \textbf{Inference inefficiency}.
Including all tool documentation significantly increases prompt length, leading to higher inference latency and computational overhead. This makes real-time applications or large-scale deployments more costly and less efficient.

Humans, by contrast, are capable of internalizing tool knowledge into their brains and applying it continuously to problem solving without consulting external tool manuals. Inspired by this, we {explore \textbf{T}ool-\textbf{In}ternalized \textbf{R}easoning (\textbf{TInR}) in LLMs.} As shown in Figure~\ref{fig:intro}(b), TInR enables reasoning with internalized tool knowledge rather than relying on external tool documentation, which could harmonize heterogeneous tool knowledge and facilitate effective and efficient tool utilization.
To realize TInR, the LLM must satisfy the following requirements: 1) \textbf{Tool internalization}. The LLM should internalize tool knowledge into its parameters, encompassing both diverse tool functionalities and strict usage rules (e.g., parameter constraints, tool call formats).
2) \textbf{Tool-reasoning coordination}. Building on internalized tool knowledge, the LLM should seamlessly integrate them into its reasoning process for adaptive tool use and strategic problem solving.


In this work, we propose \textbf{\framework}, a \textbf{T}ool-\textbf{In}ternalized \textbf{R}easoning framework with \textbf{U}nified tool usage and reasoning.
To progressively endow the LLM with TInR capability, \framework adopts a three-phase training pipeline: 
1) \textbf{Tool internalization}. The LLM is trained through a bidirectional knowledge alignment strategy, 
where it learns to map tool documentation to unique tool tokens and, conversely, to recall the original documentation from each token. We also conduct tool usage training for practical knowledge application. This design aims at both fine-grained preservation and a holistic understanding of the tool knowledge.
2) \textbf{TInR SFT warm-up}. 
We construct annotated reasoning data via rejection sampling and data formatting, ensuring both high reliability and close alignment with the TInR task.
We then employ supervised fine-tuning to equip the LLM with a foundational ability to leverage internalized tool knowledge during reasoning.
3) \textbf{TInR RL}. We employ reinforcement learning with specially designed rewards on tool tokens to further encourage exploratory tool reasoning, thereby promoting a deeper integration of tool usage competence with inherent reasoning ability. 

To comprehensively evaluate TInR capabilities, we conduct experiments in both in-domain and out-of-domain settings.
Experimental results demonstrate that our approach achieves state-of-the-art performance across both domains, showing relative improvement of $18.13\%$ in out-of-domain tool calling.

In summary, our contributions are as follows:
\begin{itemize}
\item 
We explore tool-internalized reasoning (TInR) in LLMs, aiming to facilitate reasoning with internalized tool knowledge instead of external tool documentation. 

\item 
We introduce \framework, a framework that achieves tool internalization through dedicated tool tokens, and further unifies tool usage with reasoning through a carefully designed three-phase training process.


\item 
Experiments demonstrate that \framework outperforms baselines in both in-domain and out-of-domain settings, achieving effective and efficient TInR capabilities.
\end{itemize}


\section{Related Work}

\subsection{Tool-Integrated Reasoning}
Tool-Integrated Reasoning (TIR) has recently been recognized as an effective way to enhance the reasoning capabilities of LLMs by enabling interaction with external tools. Early efforts in this direction have mainly depended on either in-context demonstrations, which guide LLMs to perform tool reasoning directly through carefully designed prompts without training~\cite{Search-o1,lu2025octotools,wu-etal-2025-agentic}, or supervised fine-tuning (SFT), which transfers tool-usage ability to smaller LLMs by distilling trajectories from stronger ones~\cite{gou2024tora,li2025startselftaughtreasonertools,chen2025learning}. However, these approaches struggle to generalize to novel tasks or unfamiliar tool settings. To overcome this limitation, more recent studies employ reinforcement learning (RL), encouraging more flexible and exploratory tool usage behaviors~\cite{li2025torlscalingtoolintegratedrl,jin2025searchr1trainingllmsreason}.
Despite recent progress, existing TIR methods still rely on external tool knowledge with limited tool understanding and efficiency. 
In this work, we explore TInR which internalizes external tool knowledge into the LLM parameters to enhance tool reasoning.

\subsection{Tool Learning in LLMs}
Tool learning aims to augment LLMs with external tools to extend their ability beyond text generation.
To achieve this, prior work has explored in-context learning, where LLMs interact directly with tool documentation in context~\cite{lumer2024toolshedscaletoolequippedagents,li-etal-2024-large-language-models,liu2025toolplanner}, as well as fine-tuning approaches that specialize LLMs on curated tool-use datasets~\cite{NEURIPS2023_8fd1a81c,tang2023toolalpaca,chen2024advancing,xu-etal-2025-petoolllm,chen2025learning}.
To address the tool size constraints in context, tool retrieval has been adopted as an upstream component to narrow down tool candidates before usage~\cite{qin2024toolllm,xu-etal-2024-enhancing-tool,shi-etal-2025-retrieval}. 
\xqc{
Recently, several studies \cite{NEURIPS2023_8fd1a81c,wang2025toolgen,su-etal-2025-toolscaler} have explored internalizing tools into LLMs.
However, these efforts are limited by small toolsets, simple reasoning strategies or unstable LLM-based evaluation \cite{iskander-etal-2024-quality}.
In contrast, our work provides a comprehensive investigation and rigorous evaluation of TInR in complex tool-use environments with large toolsets, and deliberately designs a three-phase training pipeline that enables both effective tool knowledge internalization and strategic tool-coordinated reasoning.
}

\begin{figure*}[t]
    \centering
    \includegraphics[width=1.0\textwidth]{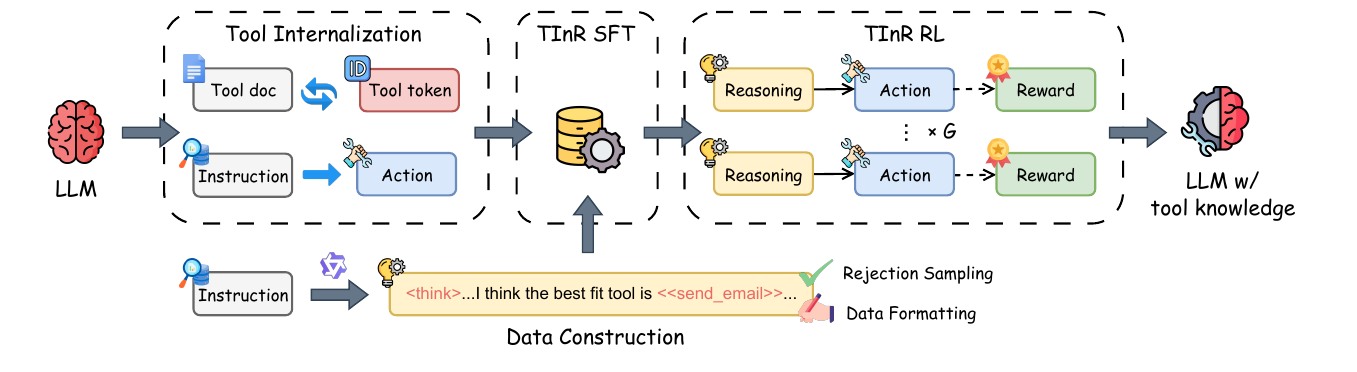}
    \caption{Illustration of our proposed \framework, with a three-phase training pipeline including tool internalization, TInR SFT warm-up, and TInR RL. \framework facilitates tool knowledge internalization and tool usage during reasoning.}
    \label{fig:main}
\vspace{-0em}
\end{figure*}

\section{Methodology}
\subsection{Task formulation}  %
Given a user instruction $q$, the goal of TInR is to solve the task through a sequence of reasoning steps interleaved with tool invocations, but without relying on any external tool documentation.
Formally, consider a tool set $\mathcal{T}=\{t_1, t_2, ..., t_N\}$, 
where 
each tool $t_i$ is associated with documentation $D(t_i)$. 
At step $j$, the LLM first performs natural language reasoning $r_j$. If a tool usage is required, the LLM executes a tool-use action $a_j$, which specifies a set of tool calls, each defined as a pair $(t, p)$ consisting of a selected tool \(t \in \mathcal{T}\) and its associated parameters $p$. The tool's output $o_j$ is then incorporated into the next reasoning step.
The overall reasoning trajectory can thus be described as:
\[
\tau = ( r_1, a_1, o_1 ), ( r_2, a_2, o_2 ), \ldots, ( r_T, a_T, o_T ).
\]
Note that the underlying decision mechanism of TInR fundamentally differs from that of TIR: tool actions in TInR are generated from internalized knowledge within the LLM rather than from external documentation.




\subsection{Overview}
\framework is a unified framework that internalizes external tool knowledge into the LLM parameters and integrates it into the reasoning process, as illustrated in Figure~\ref{fig:main}. It addresses two central requirements: 1) encoding tool functionalities and usage constraints into the LLM parameters, and 2) coordinating this internalized knowledge with multi-step reasoning to guide tool selection and invocation. 
The framework is composed of three training phases: 1)~\textbf{Phase 1} - Tool knowledge internalization. Tool functionalities and usage semantics are embedded into the LLM parameters to support tool understanding and invocation. 2)~\textbf{Phase 2} - TInR SFT warm-up. The LLM is trained with curated reasoning trajectories to align generation behavior with expected tool usage. 3)~\textbf{Phase 3} - TInR RL. Reward-driven optimization refines the robustness and accuracy of tool reasoning.


\subsection{Tool Internalization}

The first phase embeds tool knowledge into the LLM’s parameters. It consists of two steps: expanding the LLM vocabulary with dedicated tool tokens to unify reasoning and tool invocation, and aligning tool semantics and usage knowledge through a bidirectional learning objective.

\paragraph{Tool tokenization.}
To support seamless integration of tool usage within the language modeling process, the vocabulary is extended with tool-specific tokens. Each tool $\{t_i\}$ is assigned a unique token $\{I(t_i)\}$, enabling the LLM to reference and invoke tools through the next-token generation.

To reduce the action space and improve reasoning reliability, two control tags are introduced: <tool\_token> and <tool\_call>. This two-step generation first predicts a set of tool tokens $\{I(t_i)\}_{i=1}^K$ within the <tool\_token> scope, corresponding to tools $\{t_i\}_{i=1}^K \subseteq \mathcal{T}$.
Based on the associated documentation $\{D(t_i)\}_{i=1}^K$, the LLM then pairs each identified tool token with its parameters, thereby generating complete tool calls $(I(t), p)$ within the <tool\_call> scope.
We empirically demonstrate that this two-step design can benefit the TInR performance in Section~\ref{sec:ablation}.


\paragraph{Bidirectional knowledge alignment.}
After tokenization, we internalize tool knowledge within the LLM. 
Specifically, we design a bidirectional knowledge alignment strategy consisting of three objectives: tool memorization, tool recall, and tool usage grounding.

The first objective is to develop semantic mappings from the documentation to tokens.
Specifically, the LLM is trained to predict each tool token $I(t)$ based on its tool documentation $D(t)$.
Formally, the loss is defined as:
\begin{equation}
\mathcal{L}_\text{memorization} = - \sum_{t \in \mathcal{T}} \log P(I(t) \mid D(t)).
\end{equation}

The tool recall objective is to encourage fine-grained preservation of tool documentation in the internalized representation. 
Specifically, the LLM is enforced to reconstruct the original documentation $D(t)$ based on each tool token $I(t)$, which can be formulated as:
\begin{equation}
\mathcal{L}_\text{recall} = - \sum_{t \in \mathcal{T}} \sum_{s=1}^{|D(t)|} \log P ( D(t)_s \mid I(t), D(t)_{<s}), 
\end{equation}
where $D(t)_s$ denotes the $s$-th token of the documentation.


To better align tool tokens with real usage scenarios, we conduct direct tool usage training. Given each user instruction $q$ in the training dataset $\mathcal{D}$, the LLM is trained to directly generate the correct tool-use action $a$. Note that 
$a$ may consist of one or multiple tool actions, allowing the LLM to learn complex task–tool associations. The training loss is defined as: 
\begin{equation}
    \mathcal{L}_{\text{usage}} = - \sum_{q,a \in \mathcal{D}} \sum_{s=1}^{|a|} \log P ( a_s \mid q, a_{<s}).    
\end{equation}

The overall training objective is:
\begin{equation}
\mathcal{L}_\text{Phase 1} = \mathcal{L}_\text{memorization} + \alpha \mathcal{L}_\text{recall} + \beta\mathcal{L}_{\text{usage}},
\end{equation}
where $\alpha$ and $\beta$ are weighting factors.
This design enables the LLM to capture both fine-grained details and high-level semantic understanding of tool knowledge, thereby laying a solid foundation for the subsequent reasoning-oriented phases.

\subsection{TInR SFT Warm-up}
The second phase aligns the LLM’s reasoning behavior with expected tool usage through supervised fine-tuning.
\paragraph{Data Construction.}
We construct high-quality TInR trajectories via rejection sampling. Specifically, for each user instruction $q$, we collect 10 candidates tools from $\mathcal{T}$, including ground-truth, retrieved and randomly sampled tools.
Based on $q$ and the documentation of candidate tools, we prompt a large reasoning model (LRM) to synthesize multiple reasoning trajectories and only keep the correct ones validated by ground-truth tool-use actions.
To enhance tool-reasoning coordination, we further conduct data formatting by replacing each tool name field appearing in the reasoning content with its corresponding tool token, thus enabling LLMs to explicitly incorporate internalized tool knowledge into reasoning steps.
In this manner, we obtain an SFT dataset $\mathcal{D}_{\text{SFT}} = \{(q, \tau)\}$ that is highly reliable and well-aligned with the TInR goal.

We then optimize the LLM under the SFT objective:
\begin{equation}
\mathcal{L}_\text{Phase 2} = - \sum_{q,\tau \in \mathcal{D_{\text{SFT}}}} \sum_{s=1}^{|\tau|} \log P(\tau_s \mid q, \tau_{<s}),   
\end{equation}
where $\tau_s$ denotes the $s$-th token in the reasoning trajectory.


\subsection{TInR RL}
The final phase improves the robustness and adaptability of tool reasoning using reinforcement learning. A composite reward function encourages both structural correctness and accurate tool usage.
\paragraph{Reward Design.}
Rule-based reward mechanisms have demonstrated strong empirical performance and are commonly adopted in TIR methods~\cite{li2025torlscalingtoolintegratedrl,jin2025searchr1trainingllmsreason}. Following this line, we design a composite reward that combines format reward and correctness reward to ensure both structural validity and correctness of TInR trajectories.
The format reward ${R}_{\text{format}}$ verifies whether the predicted trajectory $\tau$ conforms to the required structure, i.e., contains all special tags in the correct order:
\begin{equation}
{R}_{\text{format}} = 
\begin{cases}
1, & \text{if the format of $\tau$ is correct}\\
0, & \text{otherwise}
\end{cases}
\end{equation}
The correctness reward measures both tool identification and parameter specification accuracy of tool calls.
Both accuracy is measured by the Jaccard similarity. Let $\mathcal{C}$ and $\hat{\mathcal{C}}$ denote the ground-truth and predicted tool calls. The tool reward $r_{\text{tool}}$, parameter reward $r_{\text{param}}$ and correctness reward ${R}_{\text{correct}}$ can be defined as:
\begin{equation}
r_{\text{tool}} = \left|\frac{\mathcal{I} \cap \hat{\mathcal{I}}}{\mathcal{I} \cup \hat{\mathcal{I}}}\right| \in [0, 1],
\end{equation}
\begin{equation}
r_{\text{param}} = \frac{1}{|\mathcal{C}|} \sum_{\mathcal{P}_i \in \mathcal{C}} \left|\frac{\mathcal{P}_i \cap \hat{\mathcal{P}}_i}{\mathcal{P}_i \cup \hat{\mathcal{P}}_i}\right| \in [0, 1],
\end{equation}
\begin{equation}
R_{\text{correct}} = r_{\text{tool}} + r_{\text{param}},
\end{equation}
where \( \mathcal{I} \) and \( \hat{\mathcal{I}} \) are the sets of tool tokens extracted from $\mathcal{C}$ and $\hat{\mathcal{C}}$, while $\mathcal{P}_i$ and $\hat{\mathcal{P}}_i$ denote the parameter sets of the $i$-th tool call in $\mathcal{C}$ and $\hat{\mathcal{C}}$, respectively.
The final reward is then calculated as:
\begin{equation}
{R} = {R}_{\text{format}} + {R}_{\text{correct}}.
\end{equation}

\paragraph{Training objective.}
We employ Group Relative Policy Optimization (GRPO) to optimize LLM under ${R}$.
Specifically, for each user instruction $q$, the LLM samples a group of $G$ trajectories $\{\tau_{i}\}_{i=1}^G$, where each $\tau_{i}$ is assigned a reward ${R}_i$. 
By normalizing the rewards within the group, the advantage function for $\tau_{i}$ is calculated as $A_i = \frac{{R}_{i} - \text{mean}(\{R_j\}_{j=1}^G)}{\text{std}(\{R_j\}_{j=1}^G)}$.
Then the training objective can be defined as:
\begin{equation}
\small
\begin{aligned}
    \mathcal{L}_{\text{Phase 3}} = & \mathbb{E}
_{q\sim \mathcal{D}, \tau_ \sim \pi_\theta}
\Bigg[ \frac{1}{G} \sum_{i=1}^{G}
\min\Bigg( \frac{\pi_\theta(\tau_{i} \mid q)}{\pi_{\theta_{\text{old}}}(\tau_{i} \mid q) } A_{i}, \\
& \text{clip}\left(\frac{\pi_\theta(\tau_{i} \mid q)}{\pi_{\theta_{\text{old}}}(\tau_{i} \mid q)}, 1-\epsilon, 1+\epsilon\right) A_{i} \Bigg)\Bigg],
\end{aligned}
\end{equation}
where $\pi_\theta$ is the updated policy and $\pi_{\theta_{\text{old}}}$ is the reference policy. Following~\citet{qian2025toolrl}, we remove the KL penalty term for fast adaptation to the task-specific reward.

\subsection{Inference}
During inference, the LLM follows the prescribed format with special tags 
to conduct reasoning step by step,
until ultimately resolving the user instruction. As tool knowledge is fully embedded within the LLM parameters, no external documentation or retrieval is required, allowing efficient and scalable deployment of tool-augmented reasoning in real-world applications.

\begin{table}[tbp]
\centering
 \scalebox{0.8}{
\begin{tabular}{cccc}
\toprule
Settings & Split & \# Instructions &\# Tools\\
\midrule
\multirow{3}{*}{\makecell[c]{In-domain}} 
&Train& 2552& 2467 \\
&Test(Seen)&185& 307\\
&Test(Unseen) &580& 831\\
\midrule
Out-of-domain & &1996& 2025 \\
\bottomrule
\end{tabular}}
\caption{Statistics of the experiment datasets conducted from ToolACE~\cite{liu2025toolace}, xLAM~\cite{zhang-etal-2025-xlam}, and BFCL~\cite{patil2025the}. }
\label{dataset_statistics}
\end{table}

\begin{table*}[!t]
\centering
\resizebox{\linewidth}{!}{
\begin{tabular}{@{}l|cc|ccc|cc|ccc@{}}
\toprule
\multirow{3}{*}{\textbf{Methods}} & \multicolumn{5}{c|}{\textbf{\textsc{Seen}}} & \multicolumn{5}{c}{\textbf{\textsc{Unseen}}} \\ \cmidrule(lr){2-11}
& \multicolumn{2}{c|}{\textbf{{Tool Identification}}} & \multicolumn{3}{c|}{\textbf{{Tool Calling}}} & \multicolumn{2}{c|}{\textbf{{Tool Identification}}} & \multicolumn{3}{c}{\textbf{{Tool Calling}}} \\ \cmidrule(lr){2-11}
&EM  &F1  &EM &T. Acc &P. Acc  &EM  &F1  &EM &T. Acc &P. Acc \\ \midrule
BM25+ToolRL &$48.11$ &$54.68$ &$54.59$ &$65.41$ &$61.62$ &$42.76$ &$51.24$ &$46.38$ &$59.48$ &$51.90$ \\ 
Ada-embedding+ToolRL &$56.22$ &$61.71$ &$58.38$ &$69.19$ &$67.57$ &$51.38$ &$60.19$ &$45.69$ &$60.00$ &$51.90$ \\ 
TR-Feedback+ToolRL &$65.41$ &$72.61$ &$62.70$ &$73.51$ &$71.35$ &$59.31$ &$67.80$ &$51.38$ &$67.41$ &$57.76$ \\ \hdashline
ToolRetriever+Hammer2.1-7b &$63.78$ &$69.46$ &$36.22$ &$44.32$ &$41.08$ &$59.66$ &$68.47$ &$33.28$ &$43.45$ &$34.83$ \\ 
ToolRetriever+xLAM-7b-r &$63.78$ &$69.46$ &$48.65$ &$60.00$ &$54.05$ &$59.66$ &$68.47$ &$36.38$ &$51.21$ &$40.00$ \\ 
ToolRetriever+Qwen3-8B &$63.78$ &$69.46$ &$58.38$ &$71.89$ &$64.32$ &$59.66$ &$68.47$ &$48.45$ &$68.79$ &$52.93$ \\ 
ToolRetriever+ToolRL &$63.78$ &$69.46$ &$61.08$ &$75.14$ &$70.27$ &$59.66$ &$68.47$ &$51.72$ &$67.41$ &$59.83$ \\ \hdashline
ATU &$-$ &$-$ &$61.08$ &$74.59$ &$71.35$ &$-$ &$-$ &$26.38$ &$44.14$ &$41.38$ \\ 
ToolGen &$83.78$ &$86.76$ &$71.89$ &$83.78$ &\bf 77.30 &$73.79$ &$80.55$ &$55.86$ &$72.59$ &$60.52$ \\ 
\midrule
\textbf{\framework} &\bf 85.95  &\bf 88.38 &\bf 74.05  &\bf 84.86  &\bf 77.30  &\bf 75.86 &\bf 81.86  &\bf 57.24 &\bf 74.83 &\bf 62.76 \\ 
\% improve &2.59\% &1.87\% &3.00\% &1.29\% &0.00\% &2.81\% &1.63\% &2.47\% &3.09\% &3.70\% \\
\bottomrule
\end{tabular}} 
\caption{In-domain evaluation results of baselines and \framework. EM, T.Acc, and P. Acc stand for Exact Match, Tool Accuracy, Parameter Accuracy, respectively.
\% improve represents the relative improvement achieved by our method over the previously best-performing method.}
\label{main_results}
\end{table*}
\begin{table*}[!t]
\centering
\small
\begin{tabular}{@{}l|cc|ccc@{}}
\toprule
\multirow{2}{*}{\textbf{Methods}} & \multicolumn{2}{c|}{\textbf{{Tool Identification}}} & \multicolumn{3}{c}{\textbf{{Tool Calling}}} \\ \cmidrule(lr){2-6}
&EM  &F1  &EM &T. Acc &P. Acc \\ \midrule
BM25+ToolRL &$24.94$ &$25.80$ &$16.10$ &$32.55$ &$26.46$ \\ 
Ada-embedding+ToolRL &$32.32$ &$32.38$ &$16.04$ &$36.82$ &$28.92$ \\ 
TR-Feedback+ToolRL &$30.33$ &$30.43$ &$16.39$ &$35.71$ &$29.63$ \\ \hdashline
ToolRetriever+xLAM-7b-r &$30.56$ &$30.66$ &$10.13$ &$27.17$ &$21.55$ \\ 
ToolRetriever+Hammer2.1-7b &$30.56$ &$30.66$ &$12.06$ &$28.57$ &$22.72$ \\ 
ToolRetriever+Qwen3-8B &$30.56$ &$30.66$ &$17.45$ &$36.42$ &$29.27$ \\ 
ToolRetriever+ToolRL &$30.56$ &$30.66$ &$16.63$ &\bf 37.35 &$28.45$ \\ \hdashline
ATU &$-$ &$-$ &$11.59$ &$25.29$ &$33.37$ \\ 
ToolGen &$34.89$ &$34.97$ &$22.01$ &$30.91$ &$48.24$ \\ 
\midrule
\textbf{\framework} &\bf 38.06 &\bf 38.06 &\bf 26.00 & $35.83$ &\bf 50.12 \\ 
\% improve &9.09\% &8.84\% &18.13\% &-4.07\% &3.90\% \\
\bottomrule
\end{tabular}
\caption{Out-of-domain evaluation results of baselines and \framework.} 
\label{ood}
\end{table*}

\section{Experiments}
\subsection{Setup}
\paragraph{Datasets.}
\xqc{To reflect the diversity and complexity of real-world tool environments, we conduct our experiments on three datasets, ToolACE~\cite{liu2025toolace}, xLAM~\cite{zhang-etal-2025-xlam}, and BFCL~\cite{patil2025the}, covering 1) multiple domains with large toolsets and verifiable answers, 2) both single-turn and multi-turn tasks, and 3) both in-domain and out-of-domain settings.}
For the in-domain setting, we adopt ToolACE and xLAM, where each dataset is sampled and split into training and test sets. The test set is further partitioned into seen and unseen subsets, depending on whether the ground-truth test tools appear in the training data. For the out-of-domain setting, we use 
BFCL as the test set. The statistics of datasets are summarized in Table~\ref{dataset_statistics}. 

\paragraph{Metrics.} 
We evaluate the TInR capability from two complementary dimensions: 1) Tool identification. 
This dimension measures the ability to correctly identify which tools should be used at each step. We adopt Exact Match (EM) and F1 score as evaluation metrics. 
EM captures whether the predicted set of tool tokens exactly matches the ground truth, while F1 provides a more nuanced measure by balancing precision and recall in partial matches.
2) Tool calling. To measure the ability to generate accurate tool-use actions with appropriate parameters, we use three evaluation metrics: Exact Match, which checks whether the predicted tool calls are entirely matched; Tool Accuracy, which assesses whether all tool tokens in tool calls are correct; and Parameter Accuracy, which evaluates the correctness of the predicted parameters in tool calls.



\paragraph{Baselines.} 
To provide a thorough comparison, we evaluate tool retrieval, tool reasoning and end-to-end methods.
For \textbf{tool retrieval}, we include four representative methods:
1) BM25~\cite{10.1561/1500000019}: the classical sparse retrieval method; 2) Ada Embedding: the OpenAI’s text-embedding-ada-002 model; 3) ToolRetriever~\cite{qin2024toolllm}: a dense retrieval method finetuned on tool retrieval tasks. 4) TR-Feedback~\cite{xu-etal-2024-enhancing-tool}: a dense retrieval method leveraging LLMs' iterative feedback. 
For \textbf{tool reasoning} methods, we include: 
1) Hammer-2.1-7B~\cite{lin2025robust}: a model fine-tuned with robust function-calling optimization; 
2) xLAM-7B-r~\cite{zhang-etal-2025-xlam}: a model tailored for tool usage with reasoning and action decomposition; 
3) Qwen3-8B~\cite{yang2025qwen3technicalreport}: an LRM with strong reasoning ability and built-in tool calling support.
4) ToolRL~\cite{qian2025toolrl}: an RL-based tool usage model optimized with structured rewards and Group Relative Policy Optimization (GRPO).
For \textbf{end-to-end} methods, we include:
1) ATU~\cite{li-etal-2024-towards}: an end-to-end method for direct tool usage without tool documentation.
2) ToolGen~\cite{wang2025toolgen}: a unified generation framework for both tool retrieval and tool calling using virtual tokens.
To thoroughly evaluate models across the entire pipeline, we employ ToolRL as the downstream tool usage model for each retrieval model, and ToolRetriever as the upstream tool retriever for each tool usage model.



\subsection{Main Results}
\paragraph{In-domain evaluation.}
The in-domain evaluation results are shown in Table~\ref{main_results}. 
From the results, we summarize the following key findings:
1) All baselines consistently perform worse on the unseen test set than on the seen set. This highlights the intrinsic difficulty of generalizing tool knowledge to novel tools.
2) Both tool retrieval and tool usage methods perform substantially worse than tool-internalized approaches (e.g., ToolGen). This supports our claim that it is difficult for LLMs to master tool knowledge solely from external tool documentation. 
3) Tool usage models that demonstrate strong reasoning ability (e.g., ToolRL) suffer from performance ceilings due to the upstream retrieval quality. This confirms our claim that addressing the context-length limitation of TIR through retrieval strategies is suboptimal.
4) Our proposed \framework achieves the best performance on both seen and unseen test sets with slightly larger improvements on unseen tools, demonstrating its effectiveness and generalization ability. 

\paragraph{Out-of-domain evaluation.}
We further test all methods in the out-of-domain setting, and the experimental results are shown in Table~\ref{ood}. We could observe that the performance across all methods is worse than in-domain settings, indicating that the task in this setting is more challenging.
In contrast, our method continues to achieve the best results across most metrics, with larger relative improvements on several key metrics (e.g., +18.13\% on EM for tool calling), further confirming its generalization ability.

\begin{table}[!t]
\centering
\resizebox{\linewidth}{!}{
\begin{tabular}{@{}l|cc|ccc@{}}
\toprule
\multirow{2}{*}{\textbf{Methods}} & \multicolumn{2}{c|}{\textbf{{Tool Identification}}} & \multicolumn{3}{c}{\textbf{{Tool Calling}}} \\ \cmidrule(lr){2-6}
&EM  &F1  &EM &T. Acc &P. Acc \\ \midrule
    \textbf{\framework} &\bf 78.30 &\bf 83.44 &\bf 61.31 &\bf 77.25 &\bf 66.27 \\ 
    \textit{w/o BKA} &$49.67$ &$56.81$ &$40.39$ &$48.63$ &$49.15$ \\ 
    \textit{w/o RL} &$76.47$ &$81.46$ &$59.61$ &$74.90$ &$64.18$ \\ 
    \textit{w/o recall}  &$76.21$ &$82.05$ &$59.74$ &$75.29$ &$64.58$ \\ 
    \textit{w/o memorization} &$58.43$ &$65.45$ &$45.49$ &$57.25$ &$54.64$ \\ %
    \textit{w/o usage} &$59.35$ &$66.76$ &$43.79$ &$58.56$ &$51.76$ \\ 
    \textit{w/o two-step} &$-$ &$-$ &$43.40$ &$72.94$ &$50.72$  \\ 
\bottomrule
\end{tabular}
} 
\caption{Ablation study on key components of \framework. BKA stands for bidirectional knowledge alignment.} 
\label{ablation}
\vspace{-1em}
\end{table}

\subsection{Ablation Study}
\label{sec:ablation}
We conducted an ablation study to assess the contribution of different components in our framework. First, we remove the bidirectional knowledge alignment strategy and RL training to evaluate their effect.
Then, we separately remove the three objectives in tool knowledge internalization, i.e., tool memorization, tool recall, and tool usage, to measure their individual impact. 
We also ablate the two-step design in the tool-use action to assess its importance by enforcing single-step tool calling, where LLM generates the complete tool calls directly without associating intermediate tool tokens to documentation.
Since our experiments reveal that LLMs trained without SFT warm-up are incapable of performing effective reasoning, we do not include this ablation.
Table~\ref{ablation} reports the test results in the in-domain setting.
We observe that removing bidirectional knowledge alignment strategy and RL leads to consistent performance drops, indicating their necessity for tool knowledge acquisition and strategic tool-reasoning.
All three objectives in the tool internalization phase 
(memorization, recall, usage) 
prove beneficial to performance, showing the importance of jointly preserving fine-grained tool details and grounding them in usage.
Finally, eliminating the two-step design results in substantial performance degradation, validating its efficacy in reducing the LLM’s burden to enhance reasoning. 

\subsection{In-depth Analysis}
\begin{figure}[!t]
    \centering
    \centerline{\includegraphics[width=1.0\columnwidth]{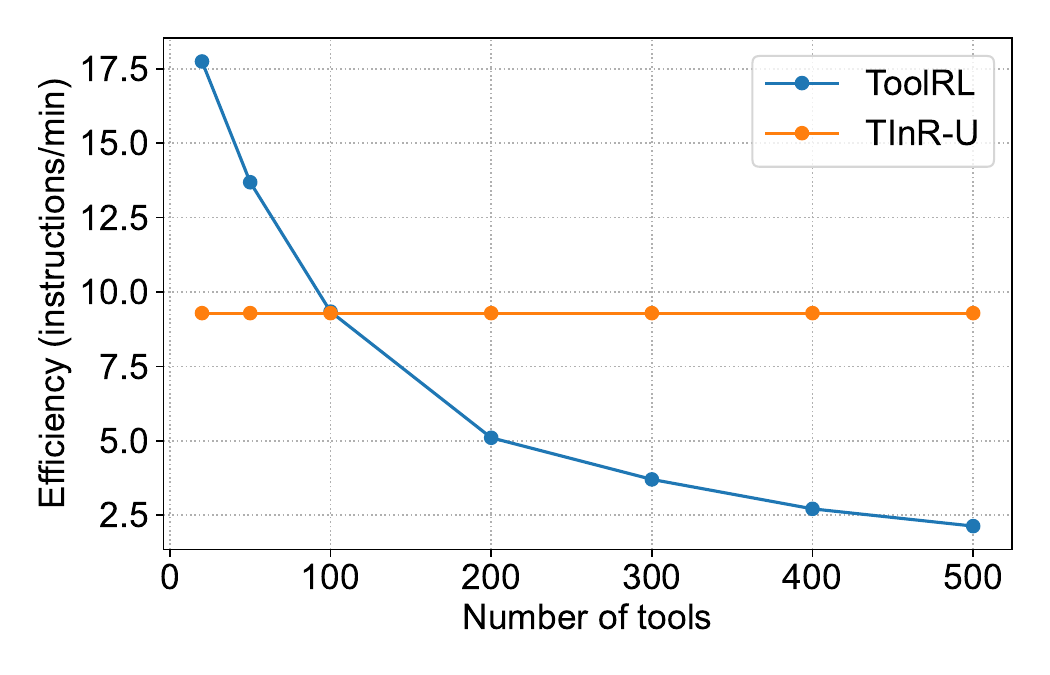}}
    \caption{Comparison of inference efficiency of ToolRL and \framework under varying tool set sizes, measured in terms of instructions processed per minute. As the tool size increases, \framework demonstrates superior efficiency.}
    \label{fig:efficiency}
\vspace{-1em}
\end{figure}

\paragraph{Analysis on inference efficiency.}
To investigate inference efficiency, we compare ToolRL, a representative TIR method, with our proposed \framework. We measure the number of user instructions each model can process per minute under varying tool set sizes, as shown in Figure~\ref{fig:efficiency}. We can observe that ToolRL’s inference speed consistently declines as the number of tools increases, which is due to the longer prompts required to append tool documentation and the resulting computational overhead. In contrast, \framework maintains constant efficiency, since tool knowledge is already internalized in the model parameters without the need for extra prompt expansion. Notably, once the tool set size exceeds 100, TInR-U surpasses ToolRL and the efficiency gap widens with larger tool sizes. This demonstrates that our approach is well-suited for real-world scenarios with numerous tools.

\paragraph{Analysis on base models.}
We further examine the robustness of \framework across different backbone LLMs. Specifically, we substitute our base model Qwen-2.5B-Instruct with LLaMA-3.1-8B-Instruct~\cite{dubey2024llama} and Mistral-7B-Instruct-v0.3~\cite{jiang2023mistral}. As shown in Table~\ref{base_model}, while absolute performance varies with model capacity and architecture, \framework consistently outperforms ToolGen under all base models,
demonstrating that our method is model-agnostic and can generalize effectively across different LLM backbones. Besides, Qwen-2.5B-Instruct achieves the strongest overall results compared to alternatives, suggesting that Qwen is more suitable as the base model for tool-internalized reasoning, likely due to its stronger built-in support for instruction following and tool usage.

\begin{table}[!t]
\centering
\resizebox{\linewidth}{!}{
\begin{tabular}{@{}l|cc|ccc@{}}
\toprule
\multirow{2}{*}{\textbf{Methods}} & \multicolumn{2}{c|}{\textbf{{Tool Identification}}} & \multicolumn{3}{c}{\textbf{{Tool Calling}}} \\ \cmidrule(lr){2-6}
&EM  &F1  &EM &T. Acc &P. Acc \\ \midrule
ToolGen (Qwen) &$76.21$ &$82.05$ &$59.74$ &$75.29$ &$64.58$ \\ 
\bf \framework (Qwen) & \bf 78.30 & \bf 83.44 & \bf 61.31 & \bf 77.25 & \bf 66.27 \\ 
\midrule
ToolGen (LLaMA) &$70.59$ &$76.44$ &$51.37$ &$67.58$ &$57.12$ \\ 
\bf \framework (LLaMA) & \bf 75.03 & \bf 80.32 & \bf 56.60 & \bf 73.33 & \bf 61.57 \\ 
\midrule
ToolGen (Mistral) &$70.46$ &$77.97$ &$51.90$ &$69.14$ &$57.07$ \\ 
\bf \framework (Mistral) & \bf 73.20 & \bf 79.18 & \bf 55.56 & \bf 72.68 & \bf 60.78 \\ 
\bottomrule
\end{tabular}
} 
\caption{Analysis of \framework on different base models.} 
\label{base_model}
\end{table}
\begin{table}[!t]
\centering
\resizebox{0.95\linewidth}{!}{
\begin{tabular}{@{}l|cc|ccc@{}}
\toprule
\multirow{2}{*}{\textbf{Methods}} & \multicolumn{2}{c|}{\textbf{{Tool Identification}}} & \multicolumn{3}{c}{\textbf{{Tool Calling}}} \\ \cmidrule(lr){2-6}
&EM  &F1  &EM &T. Acc &P. Acc \\ \midrule
    Numeric & $42.59$ & $49.83$ & $16.38$ & $41.03$ & $30.00$ \\ 
    Hierarchical & $46.72$ & $52.24$ & $19.14$ & $43.14$ & $32.07$ \\ 
    Semantic & $50.07$ & $56.32$ & $20.78$ & $48.10$ & $33.20$ \\ 
    \bf \framework &\bf 78.30 &\bf 83.44 &\bf 61.31 &\bf 77.25 &\bf 66.27 \\ 
\bottomrule
\end{tabular}
} 
\caption{Analysis on different internalization methods.}
\label{indexing}
\vspace{-1em}
\end{table}
\begin{table}[!t]
\centering
\resizebox{0.9\linewidth}{!}{
\begin{tabular}{@{}l|cc@{}}
\toprule
\textbf{Methods} & \textbf{Tool Identification} & \textbf{Tool Calling} \\ 
\midrule
DeepAgent &$62.35$ &$55.82$ \\ 
ToolRetriever+ToolRL &$60.78$ &$53.99$ \\ 
\textbf{\framework} &\bf 78.30  &\bf 61.31 \\ 
\bottomrule
\end{tabular}
} 
\caption{\xqc{Comparison with DeepAgent.}} 
\label{agent_compare}
\end{table}

\paragraph{Analysis on tool internalization methods.}
Our tool internalization method is built upon the atomic indexing strategy~\cite{NEURIPS2023_8fd1a81c,li-etal-2024-generative,wang2025toolgen}, where each tool is assigned a dedicated token for internalization. To better understand its effectiveness, we compare it against three alternative internalization approaches: 
1) Semantic indexing, which directly uses the tool name as its identifier, thereby relying on surface-level lexical semantics; 2) Numeric indexing, which assigns each tool a unique number, introducing a simple but semantically uninformative mapping; and 3) Hierarchical indexing, which clusters tools into a tree structure based on the semantic similarity of their documentation, and then assigns each tool a numerical path string from the root to its leaf.
As shown in Table~\ref{indexing}, our method achieves a large margin over all alternatives. 
These results demonstrate that our method provides unique and unambiguous representations that are easier for LLMs to memorize and recall, thereby greatly reducing confusion in tool identification and improving downstream tool calling accuracy.

\paragraph{Comparison with agent-style framework.}
We added an additional comparison with a recent agent-style framework DeepAgent~\cite{10.1145/3774904.3792460}, which equips an LLM with iterative reasoning and a scalable tool-search mechanism to select appropriate tools from large toolsets. Results are shown in Table~\ref{agent_compare}. These results suggest that DeepAgent improves over a traditional separated retrieval baseline (ToolRetriever+ToolRL), but still falls short of TInR-U. Moreover, agent-style tool retrieval typically incurs additional latency, e.g., “thinking time” before searching tools, which further reduces efficiency compared to both our approach and conventional separated retrieval pipelines.

\paragraph{Analysis on multi-turn tool-use.}
To explicitly demonstrate our method's performance on multi-step and multi-turn tool use scenarios, we separately report results on the multi-step/multi-turn subset of ToolACE and the multi-turn category of BFCL. The results are reported in Table~\ref{multi-turn}.
From the results, we can see that our \framework consistently outperforms the strongest baseline ToolGen, indicating that our gains are not limited to single-turn tool calling.

\begin{table}[!t]
\centering
\resizebox{0.9\linewidth}{!}{
\begin{tabular}{@{}l|cc|ccc@{}}
\toprule
\multirow{2}{*}{\textbf{Methods}} & \multicolumn{2}{c|}{\textbf{{Tool Identification}}} & \multicolumn{2}{c}{\textbf{{Tool Calling}}} \\ \cmidrule(lr){2-5}
&ToolACE  &BFCL  &ToolACE &BFCL \\ \midrule
    ToolGen &$73.97$  &$31.03$ &$58.90$ &$22.98$ \\ 
    \textbf{\framework} &\bf 75.34  &\bf 34.48 &\bf 61.64 &\bf 24.13 \\ 
   
\bottomrule
\end{tabular}
} 
\caption{Multi-turn tool-use results on ToolACE and BFCL datasets.} 
\label{multi-turn}
\vspace{-1em}
\end{table}
\begin{table}[!t]
\centering
\resizebox{0.5\linewidth}{!}{
\begin{tabular}{@{}l|cc@{}}
\toprule
\textbf{Methods} & \textbf{Task 1} & \textbf{Task 2} \\ \cmidrule(lr){2-3}
&EM   &EM \\ \midrule
$\theta_1$ &$60.55$ &$15.53$ \\ 
$\theta_2$ &$58.29$ &$60.52$ \\ 
\bottomrule
\end{tabular}
} 
\caption{\xqc{Experimental results in continual learning of \framework.}} 
\label{continual_learning}
\end{table}
\paragraph{Analysis on continual learning.}
\xqc{
We further assess our method's ability to handle continual learning. 
We first randomly split our test set into two subsets treated as Task 1 and Task 2, containing 398 and 367 samples respectively. We first train our base model on Task 1 to obtain $\theta_1$. Then, we extend the vocabulary to accommodate the new tools in Task 2 and continue training on Task 2 to obtain $\theta_2$. To mitigate forgetting, we adopted a rehearsal strategy by mixing a small subset of 50 Task 1 samples during Task 2 training. After training, we evaluated the tool calling accuracy of $\theta_1$ and $\theta_2$ on both tasks. From the results shown in Table \ref{continual_learning}, we observe a mild performance decrease on Task 1 from 60.55\% to 58.29\%, but a large improvement on Task 2 from 15.53\% to 59.12\%, indicating our methods' adaptability to new tasks.
}

\xqc{
We further conducted additional experiments to assess our model in tool-update scenarios. We simulate tool updates by leveraging GPT-5 to modify names or descriptions of tools and parameters in the test set. We found that the tool calling performance remains stable from 61.31\% to 58.25\%. We attribute this robustness to our two-step TInR design, where the LLM can refer to intermediate documentation before parameter filling. 
}

For more substantial changes, we acknowledge that strict zero-shot deployment may not be feasible. However, the continual learning results suggest that modest adaptation training is sufficient to recover strong performance. Overall, this reflects an explicit trade-off: while our approach may bring some maintenance costs, it delivers substantial gains in both accuracy and inference efficiency, which we believe is worthwhile in many real deployment settings.


\xqc{
We conducted additional experiments including case study, deeper ablation and efficiency analysis in Appendix \ref{additional_experiments}.
}

\section{Conclusion and Future Work}
In this paper, we explore tool-internalized reasoning (TInR), aiming at reasoning with internalized tool knowledge without relying on tool documentation. 
We propose a novel framework \framework to endow LLMs with TInR capabilities under a three-phase training process.
Extensive experiments in both in-domain and out-of-domain settings demonstrate that \framework consistently surpasses existing baselines, demonstrating both effectiveness and efficiency.
Looking ahead, we intend to explore multi-modal scenarios involving tools for vision, speech, or robotics, which could further broaden the applicability of TInR.


\section*{Limitations}
1) The evaluation datasets may not fully capture the breadth of real-world tools. However, TInR demonstrates consistent improvements across both in-domain and out-of-domain settings, suggesting strong potential to generalize beyond the evaluated scenarios.
2) The tools in our datasets may include false negatives; for example, functionally similar tools that could in principle satisfy a user’s instruction are not labeled as valid, potentially biasing tool accuracy evaluation. However, this issue is inherent to many tool datasets and, given that such cases are infrequent, their effect on our results is likely negligible.

\section*{Ethics Statement}
The dataset used in our work is derived from publicly available sources and generated through interactions with LLMs in English. Since the SFT reasoning data in our study are entirely simulated, user privacy is fully protected, and no real personal information is included in the dataset. Furthermore, all scientific artifacts used in this research are publicly accessible for academic purposes under permissive licenses, and their use in this paper complies with their intended purposes. Given these considerations, we believe our research adheres to the ethical standards of the conference.

\section*{Acknowledgments}
The work described in this paper was supported by Research Grants Council of Hong Kong (PolyU/15207122, PolyU/15209724, PolyU/15213323, PolyU/15205325), PolyU internal grants (BDWP) and the National Natural Science Foundation of China under Grants 62476071.

\bibliography{anthology,custom}

\appendix

\section{Implementation Details}
\label{implementation_detail}
In data construction, we employ Qwen3-14B as the LRM to generate reasoning data. Our candidate tool set consists of three parts: the ground-truth tools along with 5 tools retrieved using ToolRetriever, and the remaining tools randomly sampled.
We train \framework based on Qwen2.5-7B-Instruct~\cite{yang2024qwen2}.
In Phase 1 training of \framework, we fine-tune the Qwen2.5-7B-Instruct model with a learning rate set to $5e{-5}$, a batch size of $64$ and a warm-up ratio of $0.1$, for $8$ epochs. The weighting factor $\alpha$ and $\gamma$ are set to 1.
In Phase 2 training of \framework, we fine-tune with a learning rate set to $5e{-6}$ and a batch size of $64$ for $4$ epochs. 
For reinforcement learning in Phase 3, we use GRPO with a learning rate set to $2e{-6}$ and a batch size of $128$ for $20$ epochs. 
We have trained the model several times to ensure the improvement is not randomly achieved and present the mid one. 
To accelerate the memorization of tools that without instructions, we generate pseudo-instructions for these tools.
Since the maximum context length varies in different LLMs, we constrain the context window to 4096 tokens. The experiments are conducted on NVIDIA 5880 GPUs with 48 GB of memory. 

\section{Additional Experiments and Analysis}
\label{additional_experiments}
\subsection{Case Study}
\label{case_study}
\xqc{
To further demonstrate how internalized tool knowledge manifests in reasoning behavior, we provide more qualitative evidence and deepen the ablation analysis by conducting a concrete case study comparing \framework against its variant without bidirectional knowledge alignment.
As shown in the Figure~\ref{fig:case_study}, when the user requests information about the `Fnatic` League of Legends team and the full friends list of a Steam user, TInR-U correctly grounds its reasoning on the available tool tokens (e.g., `<<get\_teams\_and\_players>>` and `<<user\_friends\_list>>` ). In contrast, the ablated model tends to hallucinate tool tokens and parameter names (e.g., inventing `\_esports` and `\_friends` tools and `teamname`parameters). This highlights that the bidirectional knowledge alignment in our framework not only improves scores but also leads to more faithful, schema-consistent reasoning traces. 
}

\definecolor{lightgray}{RGB}{240, 240, 240}
\lstdefinestyle{prompt}{
    basicstyle=\ttfamily\fontsize{7pt}{8pt}\selectfont,
    frame=none,
    breaklines=true,
    backgroundcolor=\color{lightgray},
    breakatwhitespace=true,
    breakindent=0pt,
    escapeinside={(*@}{@*)},
    numbers=none,
    numbersep=5pt,
    xleftmargin=5pt,
}
\tcbset{
  aibox/.style={
  }
}
\newtcolorbox{AIbox}[2][]{aibox, title=#2,#1}

\begin{figure*}[!ht] 
\vspace{-5mm}
\begin{AIbox}{Case Study}
{\bf User:} \\
{
Could you check for any giveaways on the Xbox One platform and also on the Android platform? \\


} 

{\bf Assistant (TInR-U):} \\
{
<think>\\
Okay, let's see. The user wants information about the 'Fnatic' team in League of Legends Esports and their players. Also, they need the full friends list of a Steam user with ID '77788899900011122'. \\
First, for the League of Legends Esports team, there's the <<get\_teams\_and\_players>> tool. The parameter is the team name, which is 'Fnatic'. So I'll use that tool with team\_name set to 'Fnatic'. \\
Next, the Steam friends list. The <<user\_friends\_list>> tool requires the steamid, which is given as '77788899900011122'... \\
</think>\\
...\\
<tool\_call>\\
\{"token": "<<get\_teams\_and\_players>>", "parameters": \{"name": "Fnatic"\}\} \\
\{"token": "<<user\_friends\_list>>", "parameters": \{"is\_id": "77788899900011122"\}\} \\
</tool\_call>\\

}

{\bf Assistant (TInR-U wo/ bidirectional knowledge alignment):} \\
{
<think>\\
Okay, let's tackle the user's request step by step. First, they want information about the 'Fnatic' team in League of Legends Esports and their players. \\
Looking at the available tools, there's the \_esports function which takes a team name. The default is 'lng-esports', but the user specified 'Fnatic', so I'll need to set the parameter to 'Fnatic'. \\
Next, the user also wants the full friends list of a Steam user with ID '77788899900011122'. The \_friends tool requires a user\_id as an integer. The provided ID is a string, but the tool accepts integers, so I'll convert it to 77788899900011122... \\
</think>\\
...\\
<tool\_call>\\
\{"token": "\_esports", "parameters": \{"teamname": "Fnatic"\}\} \\
\{"token": "\_friends", "parameters": \{"is\_id": "77788899900011122"\}\} \\
</tool\_call>\\
}

\end{AIbox} 
\caption{\xqc{A case study comparing \framework against its variant without bidirectional knowledge alignment.}}
\label{fig:case_study}
\vspace{-5mm}
\end{figure*}

\subsection{More Ablation Analysis}
\label{step_design}
\xqc{
The ablation results in Table \ref{ablation} shows that the effect of removing recall is minor compared with memorization and usage in our current two-step TInR design.
To further probe the impact of recall, we conducted an additional ablation in the single-step setting, where the LLMs generate the complete tool calls directly without associating intermediate tool tokens to documentation. The results are summarized in Table \ref{ablation_recall}. We observe that in the single-step setting, removing recall leads to substantially larger drops, especially on parameter accuracy. This indicates that when the reasoning and parameter-filling burden is higher, recall contributes more strongly to model performance. Thus, our findings suggest that the two-step architecture itself already simplifies the recall burden, and recall becomes more crucial when the model must internally manage more complex parameter inference.
}

\subsection{More Efficiency Analysis}
\label{sec:efficiency_scalability}
\xqc{
To broaden the scalability evidence in efficiency, we conducted additional experiments on (i) different hardware and (ii) a larger model scale.
We first replace the NVIDIA 5880 GPUs used in our original experiments with NVIDIA A6000 GPUs and measure efficiency across tool sets ranging from 200 to 500 tools. The results are summarized in Table \ref{efficiency_hardware}. As shown, both models run slightly slower on A6000 GPUs, likely due to architectural and memory bandwidth differences. However, TInR-U consistently maintains around 9 instructions/min on both hardware platforms and remains substantially faster than ToolRL. This confirms that the scalability advantage of our approach is robust to hardware variation.
Due to computational resource constraints to fully finetune a larger 14B model, we apply LoRA-based training on Qwen2.5-14B-Instruct and measure inference efficiency. We observe that TInR-U achieves 7.41 instructions/min, which still outperforms ToolRL. These results show that the TInR-U scales reliably across both hardware platforms and model sizes.
We further evaluate the inference efficiency of ToolGen for comparison under the same setting, and observe that ToolGen achieves 9.47 instructions/min, which is comparable to our \framework with 9.29 instructions/min. The slight improvement is likely because our model is trained with GRPO which is known to increase reasoning length; however, this is acceptable given our performance gains. Importantly, both ToolGen and \framework are substantially faster than ToolRL when the tool set exceeds 100 tools, 
indicating that TInR is essential for maintaining fast inference on large toolsets.
}

\begin{table}[!t]
\centering
\resizebox{\linewidth}{!}{
\begin{tabular}{@{}l|ccc@{}}
\toprule
\textbf{Methods} & \multicolumn{3}{c}{\textbf{{Tool Calling}}} \\ \cmidrule(lr){2-4}
&EM &T. Acc &P. Acc \\  \midrule
two-step & $61.31$ & $77.25$ & $66.27$ \\ 
\textit{w/o recall} &$59.74$(-$2.6\%$) &$75.29$(-$2.5\%$) &$64.58$(-$2.6\%$) \\ \midrule
single-step  &$43.40$ &$72.94$ &$50.72$ \\ 
\textit{w/o recall} &$40.13$(-$7.5\%$) &$70.20$(-$3.8\%$) &$46.67$(-$8.0\%$) \\
\bottomrule
\end{tabular}
} 
\caption{\xqc{Ablation of recall in two-step and single-step design of \framework on tool calling performance.}} 
\label{ablation_recall}
\end{table}
\begin{table}[!t]
\centering
\resizebox{0.8\linewidth}{!}{
\begin{tabular}{@{}l|cc@{}}
\toprule
\textbf{Methods} & \textbf{NVIDIA 5880} & \textbf{NVIDIA A6000} \\ 
\midrule
ToolRL &[$2.13,5.10$] &[$2.07,4.86$] \\ 
TInR-U &$9.29$ &$8.97$ \\ 
\bottomrule
\end{tabular}
} 
\caption{\xqc{Comparison of inference efficiency of ToolRL and \framework with toolsets ranging from 200 to 500 tools across different hardware.}}
\label{efficiency_hardware}
\end{table}
\begin{table}[!t]
\centering
\resizebox{0.9\linewidth}{!}{
\begin{tabular}{@{}l|cc@{}}
\toprule
\textbf{Methods} & \textbf{Tool Identification} & \textbf{Tool Calling} \\ \cmidrule(lr){2-3}
&EM   &EM \\ \midrule
$\alpha$=0.5,  $\beta$ =1.0 &$76.73$ &$59.74$ \\ 
$\alpha$=2.0,  $\beta$ =1.0 &$77.51$ &$60.52$ \\ 
$\alpha$=1.0,  $\beta$ =0.5 &$77.12$ &$58.69$ \\ 
$\alpha$=1.0,  $\beta$ =2.0 &$78.56$ &$61.05$ \\ 
$\alpha$=1.0,  $\beta$ =1.0 &$78.30$ &$61.31$ \\ 
\bottomrule
\end{tabular}
} 
\caption{\xqc{Sensitivity analysis of hyperparameters in \framework.}}
\label{hyperparameter}
\end{table}

\subsection{Hyperparameter Sensitivity Analysis}
\xqc{
We conducted additional sensitivity analysis on the weighting factors $\alpha$ and $\beta$ with varying values. As shown in the Table \ref{hyperparameter}, the performance of \framework remains stable across a broad range of hyperparameter settings, with fluctuations within only 1–2 points for both tool identification and tool calling accuracy. This indicates that our method does not rely on delicate tuning and that the default setting used in the paper is already near-optimal and robust.
}

\subsection{Theoretical Discussion}
\xqc{
To further elucidate the effect of bidirectional knowledge alignment, we provide a theoretical discussion of its underlying principles. Similar to many works in tool learning and representation alignment, our bidirectional knowledge alignment is motivated by its effect on representation alignment across three levels: (i) tool documentation → (ii) tool-token embeddings → (iii) the LLM's internal language space. By enforcing alignment in both directions (i.e., memorization and recall), the LLM learns to ground tool semantics in a way that is consistent for both discrimination (i.e., identifying the appropriate tool) and generation (i.e., producing accurate parameters), thereby facilitating both fine-grained preservation of tool details and holistic understandings of tool functionalities.
}

\section{Prompt Details}
The prompt template for inference are shown in Figure~\ref{fig:prompt_example}.

\begin{figure*}[!b] 
\vspace{-5mm}
\begin{AIbox}{Prompt for TInR inference}
{\bf Prompt:} \\
{
You are a helpful assistant capable of leveraging your tool knowledge to address the user's query given inside <user> and </user>.
First, you should conduct reasoning inside <think> and </think>. During reasoning, you can generate special tool tokens from your vocabulary to recall your tool knowledge. If one or more tools are needed, identify all the corresponding tool tokens inside <tool\_token> and </tool\_token>. Next, you will be given the corresponding tool documentation of all valid tool tokens. You should specify the tool token and corresponding parameters inside <tool\_call> and </tool\_call> to invoke the tool call. \\

Output Format for Tool Identification: \\
<think> Your reasoning process containing special tool tokens (if possible) </think> \\
<tool\_token> \\
Tool token
... \\
</tool\_token>\\

Output Format for Tool Calling: \\
<tool\_call> \\
{"token": "Tool token", "parameters": {"Parameter name": "Parameter value", "... ...": "... ..."}} \\
{"token": "... ...", "parameters": {"... ...": "... ...", "... ...": "... ..."}}
...\\
</tool\_call> \\
\\
Remember: \\
1. You must always include the <think>, <tool\_token> or <tool\_call> fields to outline your reasoning and then specify tool tokens or tool calls.
2. You can invoke multiple tool calls in the <tool\_call> field, where each should be specified in JSON format with a "token" field and an "parameters" field containing a dictionary of parameters. If no parameters are needed, leave the \"parameters\" field an empty dictionary.
3. Refer to the Dialogue Records History, including the user's previous queries, previous tool calls or responses noted as <tool\_call> or <response>, and any tool feedback noted as <obs> (if exists).





}

\end{AIbox} 
\caption{The prompt for TInR inference.}
\label{fig:prompt_example}
\vspace{-5mm}
\end{figure*}

\label{sec:appendix}

\end{document}